# From Keypoints to Realism: A Realistic and Accurate Virtual Try-on Network from 2D Images

Maliheh Toozandehjani, Ali Mousavi[*], Reza Taheri

Department of Computer Engineering, Ne. C., Islamic Azad University, Neyshabur, Iran
E-mails: maliheh.toozandehjani@iau.ir; mousavi@iau.ac.ir; reza.taheri@iau.ir
[*] means corresponding author

**Short Abstract**
The aim of image-based virtual try-on is to generate realistic images of individuals wearing target garments, ensuring that the pose, body shape and characteristics of the target garment are accurately preserved. Existing methods often fail to reproduce the fine details of target garments effectively and lack generalizability to new scenarios. In the proposed method, the person's initial garment is completely removed. Subsequently, a precise warping is performed using the predicted keypoints to fully align the target garment with the body structure and pose of the individual. Based on the warped garment, a body segmentation map is more accurately predicted. Then, using an alignment-aware segment normalization, the misaligned areas between the warped garment and the predicted garment region in the segmentation map are removed. Finally, the generator produces the final image with high visual quality, reconstructing the precise characteristics of the target garment, including its overall shape and texture. This approach emphasizes preserving garment characteristics and improving adaptability to various poses, providing better generalization for diverse applications.

## 1. Short Introduction (4-5 lines)

This paper analyzes and enhances the technology of image-based virtual try-on, which improves the online shopping experience by generating realistic images of individuals wearing target garment. This technology allows buyers to virtually try-on garment without worrying about size and model, offering economic benefits for retailers as well. The paper evaluates current challenges in existing methods, such as the inability to reproduce accurate details of the body and garment and the lack of generalization to new poses, and introduces an innovative approach for more precise garment warping. This method demonstrates more effective qualitative results and promises significant advancements in the field of virtual try-on.

## 2. Proposed Work and Methodology (including comprision, simulation/experimental results and discusion)

The VITON-HD+ method represents a remarkable advancement in the field of image-based virtual try-on. This approach precisely removes the initial garment and employs keypoint prediction based on pose to warp the target garment with high accuracy, ensuring the preservation of its features. Additionally, by predicting the body segmentation map, the detailed structure of the person's body is accurately reconstructed. The use of an alignment-aware segment normalization eliminates potential errors between the warped garment and the predicted garment region. Finally, the generator produces a high-resolution output image. Qualitative experiments on the VITON-HD dataset demonstrate that this proposed method excels both in preserving the features of the target garment and in adapting to new conditions and scenarios.

## 3. Conclusion (4-5 lines)

In this study, we propose VITON-HD+, a novel image-based virtual try-on network that produces realistic results. By using a keypoint prediction module for garment warping, we achieve more precise and accurate warping without needing an independent display of the person's garment. Based on the warped garment, a body part segmentation map is predicted while the person is wearing the target garment. Finally, by using a normalization method for aligned parts, the final try-on image is produced through the generator. Qualitative comparisons indicate that our approach significantly outperforms existing virtual try-on methods by preserving the complete details, shape, and texture of the target garment, and shows strong generalizability across different scenarios.

# از نقاط کلیدی تا واقع‌گرایی: یک شبکه واقع‌گرایانه و دقیق برای پُرو مجازی از تصاویر دوبعدی

**ملیحه توزنده‌جانی**
گروه مهندسی کامپیوتر، واحد نیشابور، دانشگاه آزاد اسلامی، نیشابور، ایران

**سید علی موسوی**
گروه مهندسی کامپیوتر، واحد نیشابور، دانشگاه آزاد اسلامی، نیشابور، ایران

**رضا طاهری**
گروه مهندسی کامپیوتر، واحد نیشابور، دانشگاه آزاد اسلامی، نیشابور، ایران

**چکیده**
هدف پُرو مجازی مبتنی بر تصویر، تولید تصاویر واقع‌گرایانه از اشخاص با لباس‌های هدف است، به‌گونه‌ای که ژست، شکل بدن شخص هدف و ویژگی‌های لباس هدف با دقت حفظ شوند. روش‌های موجود اغلب در بازتولید ویژگی‌های دقیق لباس هدف ناکارآمد بوده و توانایی تعمیم به موقعیت‌های جدید را ندارند. در این روش پیشنهادی، ابتدا لباس اولیه شخص به‌طور کامل حذف می‌گردد. سپس، با بهره‌گیری از نقاط کلیدی پیش‌بینی‌شده، دفرمه‌سازی دقیقی برای تطبیق کامل لباس هدف با ساختار بدن و ژست شخص انجام می‌شود. بر اساس لباس دفرمه‌شده، نقشه قطعه‌بندی بدن با دقت بالاتری پیش‌بینی شده و به کمک فرآیند نرمال‌سازی بخش‌های مرتب‌شده، نواحی نامرتب ایجاد شده بین لباس دفرمه‌شده و ناحیه لباس پیش‌بینی‌شده در نقشه قطعه‌بندی حذف می‌شوند. در نهایت، ژنراتور تصویر پرونهایی را با کیفیت بصری بالا تولید کرده و ویژگی‌های دقیق لباس هدف، از جمله شکل کلی و بافت، را بازسازی می‌نماید. این روش پیشنهادی با تأکید بر حفظ ویژگی‌های لباس و بهبود تطبیق با ژست‌های مختلف، تعمیم‌پذیری بیشتری را برای استفاده‌های گسترده و متنوع فراهم می‌کند.

**کلمات کلیدی**
پُرو مجازی، لباس دفرمه‌شده، نقشه قطعه‌بندی بدن شخص.

## 1- مقدمه

شبکه پُرو مجازی مبتنی بر تصاویر دو بعدی که به‌طور خلاصه VITON[1] نامیده می‌شود، به فرآیند تولید تصاویر واقع‌گرایانه‌ای اشاره دارد که در آن تصویری مصنوعی تولید می‌شود که شخص هدف با لباس هدف را پوشیده است. این فناوری امکان تجربه خرید آنلاین بدون نگرانی از سایز و مدل لباس را فراهم می‌کند و کاربران می‌توانند با بهره‌گیری از این فناوری، به راحتی لباس‌های مختلف را به صورت آنلاین پُرو کرده و بهترین انتخاب را داشته باشند، حتی اگر نتوانند به فروشگاه‌های فیزیکی مراجعه کنند. این فناوری به افزایش شانس خرید موفق و رضایت مشتری کمک کرده و تجربه خرید آنلاین را به‌طور قابل‌توجهی بهبود می‌بخشد. علاوه بر این، پرو مجازی برای خرده‌فروشان مزایای چشمگیری دارد. این فناوری به آن‌ها اجازه می‌دهد قیمت‌های کمتری ارائه دهند و نیز هزینه‌های بازگشت محصول به دلیل سایز نامناسب یا ظاهر غیرمنتظره را کاهش دهند. پُرو مجازی در تصویر تجربه‌های استایل شخصی و مدیریت کمد لباس را بهبود بخشیده است. همچنین، طراحان مد می‌توانند از شبکه‌های پُرو مجازی برای تجسم طرح‌های خود بر روی انواع استایل‌های بدن بدون نیاز به مدل‌های واقعی استفاده کنند که این امر فرآیند طراحی را کارآمدتر، مقرون‌به‌صرفه‌تر و خلاقانه‌تر می‌کند. شبکه‌های مبتنی بر شبکه پُرو مجازی مبتنی بر تصاویر دو بعدی را می‌توان به عنوان یک مسئله تولید تصویر با شرایط خاص در نظر گرفت. در شبکه‌های پُرو مجازی مبتنی بر تصویر، ورودی شامل تصویر شخص هدف $I$ و لباس هدف $c$ بوده و خروجی مطلوب تصویر پُرو نهایی $\hat{I}$ است بطوری‌که این تصویر مصنوعی تولید شده باید دارای معیارهایی باشد، از جمله:

✓ شکل بدن و ژست شخص در تصویر ورودی باید حفظ شود و اعضا و جزئیات ظریف بدن مانند انگشتان دست باید به وضوح ارائه شوند.

✓ لباس هدف باید به خوبی با ناحیه مربوطه بدن شخص هدف مطابقت داشته باشد و به‌صورت هموار و یکپارچه دفرمه شود.

✓ شکل کلی و تمامی جزئیات لباس هدف مانند بافت، لوگو و گلدوزی تا حد امکان باید حفظ شود.

✓ سایر اعضا بدن که نیاز به جایگزینی ندارند، باید به‌درستی حفظ شوند و کمترین تغییرات را داشته باشند. این شامل لباس‌های پایین‌تنه، سر و صورت شخص هدف می‌شود که باید بدون تغییرات عمده باقی بمانند.

اصولاً شبکه‌های پُرو مجازی مبتنی بر تصاویر دو بعدی شامل دو ماژول دفرمه‌سازی[2] و پُروکردن[3] می‌باشند. ماژول دفرمه‌سازی وظیفه دفرمه‌سازی تصویر لباس هدف را بر عهده دارد. سپس تصویر نهایی شخص هدف که لباس هدف را بر تن کرده است، توسط ماژول پُروکردن تولید می‌شود. یکی از روش‌های پایه در پُرو مجازی مبتنی بر تصویر، روش CP-VTON است [1]. همان‌طور که در شکل 1 مشاهده می‌شود، این روش جزئیات بدن و لباس هدف را به خوبی نمایش نمی‌دهد و تصاویر پُرونهایی از کیفیت خوبی برخوردار نیستند. از سوی دیگر، روش VITON-HD [2] به عنوان یکی از روش‌های پیشرفته، با پیش‌بینی نقشه قطعه‌بندی بدن، ناحیه مربوط به لباس هدف را تعیین می‌کند. سپس بر اساس این ناحیه پیش‌بینی‌شده، لباس هدف را دفرمه کرده و تصویر پُرونهایی را با توجه به نقشه قطعه‌بندی پیش‌بینی‌شده و لباس دفرمه‌شده تولید می‌کند. با این حال VITON-HD نیز قادر به حفظ جزئیات دقیق لباس هدف نیست و تصاویر پُرونهایی در نواحی یقه تحت تأثیر لباس اولیه قرار می‌گیرند (به شکل 1 مراجعه شود). علی‌رغم پیشرفت‌های انجام‌شده، تصاویر تولیدشده توسط روش‌های موجود هنوز به واقعیت نزدیک نیستند و با چالش‌های متعددی مواجه‌اند. از جمله این چالش‌ها می‌توان به ناتوانی در حذف کامل لباس‌های اولیه، عملکرد ضعیف در مقابله با سبک‌های مختلف لباس و دشواری در پُرو سایزهای متفاوت اشاره کرد. بسیاری از روش‌ها تنها شکل لباس را در نظر می‌گیرند و به اطلاعات اندازه توجهی نمی‌کنند، که این مسئله موجب کاهش دقت و کیفیت تصاویر پُرونهایی می‌شود. مطالعات نشان داده‌اند که روش‌های فعلی با مشکل از‌دست‌دادن ویژگی‌های لباس‌های دفرمه‌شده مواجه هستند و نقشه‌های قطعه‌بندی اعضا بدن به طور کامل از جمله انگشتان دست را پوشش نمی‌دهند. علاوه بر این، اکثر روش‌ها بیشتر بر روی لباس‌های بالاتنه تمرکز دارند و لباس‌های پایین‌تنه را نادیده می‌گیرند. همچنین، عدم توانایی در پُرو هم‌زمان لباس‌های بالاتنه و پایین‌تنه و محدودیت تنوع مجموعه‌داده‌ها از دیگر چالش‌های مهم به شمار می‌آیند. این محدودیت‌ها نیازمند بهبود و توسعه بیشتر در زمینه پُرو مجازی هستند.

---

[1] Image-based Virtual Try-on Network
[2] Geometric Matching Module
[3] Try-on Module

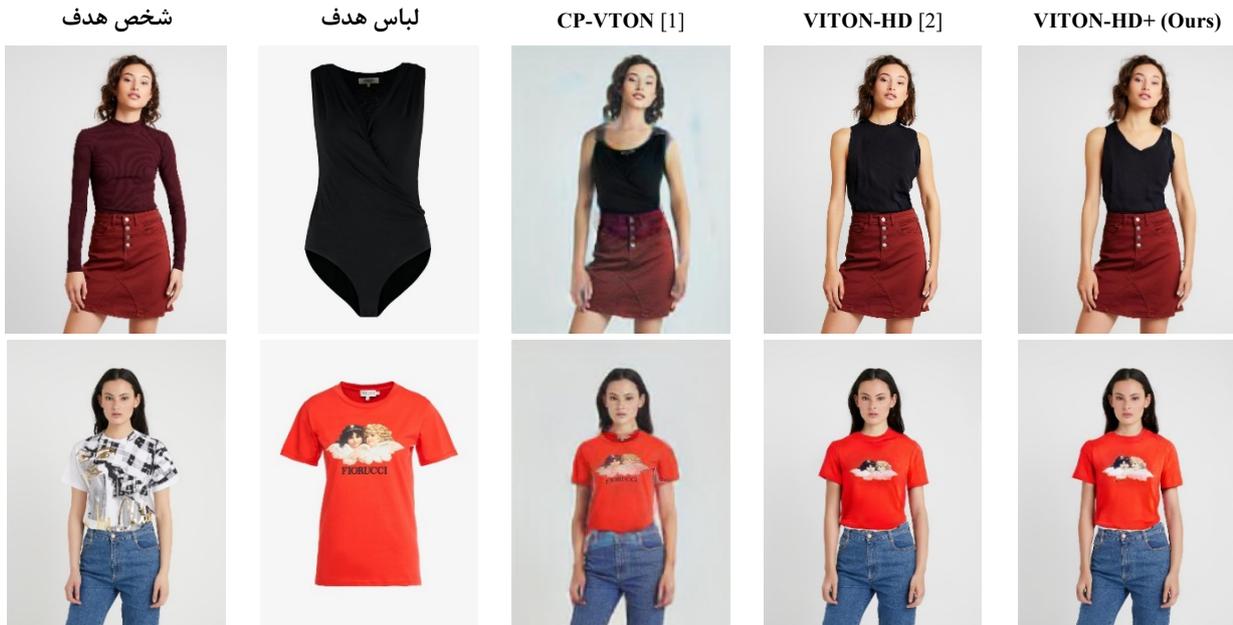

شکل ۱- از چپ به راست به ترتیب در ستون اول تصویر شخص هدف با لباس اولیه، تصویر لباس هدف، نتایج پُرونهایی تولید شده توسط روش‌های CP-VTON، VITON-HD و +VITON-HD پیشنهادی نشان داده شده است.

در این مقاله، چالش‌های اصلی شامل دقت ناکافی در تولید ویژگی‌های لباس هدف از جمله شکل کلی و بافت لباس و عدم تعمیم‌پذیری به موقعیت‌های جدید مورد بررسی قرار گرفته‌اند. برای رفع این چالش‌ها، شبکه پرو مجازی مبتنی بر تصویر به نام +VITON-HD ارائه شده است. همان‌طور که در شکل ۱ مشاهده می‌شود، این روش با نمایش مستقل از لباس شخص، لباس اولیه را به‌درستی حذف کرده و قابلیت تعمیم به موقعیت‌های مختلف را دارد. ماژول پیش‌بینی نقاط کلیدی لباس بر اساس ژست، به‌منظور کمک به دفرمه‌سازی لباس هدف با حفظ ویژگی‌ها استفاده شده است تا لباس هدف بدون وابستگی به نمایش مستقل از لباس شخص با دقت دفرمه شود. بر اساس این لباس دفرمه دقیق‌تر، نقشه قطعه‌بندی بدن شخص پیش‌بینی می‌شود و جزئیات بدن شخص به‌طور دقیق حفظ می‌شوند. با استفاده از روش نرمال‌سازی بخش‌های مرتب‌شده[4]، اطلاعات نادرست بین ناحیه لباس پیش‌بینی‌شده و لباس دفرمه‌شده حذف می‌شوند و در نهایت، تصویر پُرونهایی با کیفیت بالا توسط ژنراتور تولید می‌شود. به طور خلاصه، +VITON-HD موارد زیر را به عنوان دستاوردهای خود معرفی می‌کند:

- معرفی یک ماژول پیش‌بینی نقاط کلیدی لباس بر اساس ژست به عنوان ورودی ماژول دفرمه‌سازی، برای حفظ دقیق ویژگی‌های لباس‌های هدف و عدم وابستگی به نمایش مستقل از لباس شخص.
- توسعه یک ماژول پیش‌بینی نقشه قطعه‌بندی بر اساس لباس‌های دفرمه دقیق‌تر، که اعضا مختلف بدن را به‌خوبی متمایز می‌کند.
- آزمایشات کیفی بر اساس مجموعه‌داده VITON-HD به‌طور کامل نشان می‌دهند، این روش‌پیشنهادی به عملکرد فوق‌العاده‌ای در حفظ ویژگی‌های لباس هدف و تعمیم‌پذیری به موقعیت‌های جدید دست می‌یابد.

در بخش دوم مقاله به بررسی کارهای مرتبط و مقایسه روش‌های موجود در حوزه پرو مجازی پرداخته می‌شود. بخش سوم شامل توضیحات مفصل روش‌پیشنهادی که شامل مراحل مختلف و تکنیک‌های به‌کارفته در هر مرحله است. در بخش چهارم، ارزیابی عملکرد روش‌پیشنهادی به‌همراه مقایسه با روش‌های موجود ارائه می‌شود. به طور خاص، تمرکز بر نتایج کیفی است تا تأثیرات و عملکرد روش‌پیشنهادی به‌طور کامل بررسی و ارزیابی شود. بخش پنجم به بحث و نتیجه‌گیری اختصاص دارد، که در آن دستاوردهای اصلی مقاله و پیشنهادات برای کارهای آینده مطرح می‌شود.

## 2- کارهای مرتبط

در دنیای پرو مجازی مبتنی بر تصاویر دو بعدی، روش‌های متعددی برای بهبود تجربه کاربران و ارائه تصاویر واقعی‌تر ارائه شده‌اند. اصولاً می‌توان کلیه روش‌های رایج را به دو دسته مختلف دسته بندی کرد: - روش‌های دومرحله‌ای و - روش‌های سه‌مرحله‌ای، که در ادامه به بررسی آن‌ها می‌پردازیم.

### 1-2- روش های دومرحله ای

میان تمامی روش‌های مبتنی بر تصاویر دو بعدی، VITON [3] و CP-VTON [1] دو رویکرد پایه‌گذار هستند، که استراتژی دو مرحله‌ای شامل ماژول دفرمه‌سازی و پُرکردن را به‌کارگرفته‌اند. VITON با استفاده از پیش‌بینی تبدیل اسپلاین صفحه نازک (TPS)[5] و تطبیق زمینه‌شکل [4] لباس هدف را دفرمه می‌سازد. این روش از یک استراتژی تولید تصویر درشت به ریز برای انتقال لباس هدف به ناحیه مربوطه بدن شخص هدف استفاده می‌کند. CP-VTON به‌منظور بهبود روش VITON، یک ماژول دفرمه‌سازی پیشرفته معرفی کرده است که پارامترهای تبدیل TPS را به‌صورت انتها به انتها (مانند [5]) فرامی‌گیرد. این روش منجر به تولید نتایج پرو مجازی بهتری با حفظ بیشتر جزئیات لباس هدف شده است. SP-VITON [6] دقیقاً رویکرد مشابهی با VITON اتخاذ کرده است، با این تفاوت که ژست دو بعدی زیر لباس شخص با استفاده از DensePose [7] برای وابستگی کمتر به لباس اولیه، پیش‌بینی می‌شود. رویکردهای بعدی، ماژول دفرمه‌سازی و پُرکردن ارائه‌شده در CP-VTON را با مکانیسم‌های مختلف بهبود بخشیده‌اند تا تصاویر پُرونهایی باکیفیت‌تری تولید کنند. در ادامه CP-VTON+ [8] با اصلاح توابع زیان و نمایش مستقل از لباس شخص ارائه‌شده در VITON به عنوان ورودی، به دفرمه‌سازی دقیق‌تر لباس‌ها و تولید تصاویر با کیفیت‌تر پرداخته است. VITON-GAN [9] نیز با افزودن تابع زیان خصمانه[6] در مرحله پُرکردن، کیفیت تصاویر پُرونهایی را به‌ویژه در مواجهه با ژست‌های پیچیده (مانند بازوهایی که در جلوبدن روی هم قرارگرفته‌اند) بهبود بخشیده است. برای دستیابی به دفرمه‌سازی واقعی‌تر لباس‌ها، روش LA-VITON [10] با استفاده از یک فرآیند دو مرحله‌ای شامل تبدیل Perspective و TPS و تکنیک‌های پیشرفته‌ای مانند تابع زیان ثبات فاصله شبکه (GIC loss)[7]، کنترل انسداد (OHT)[8] و GAN loss به بهبود قابل‌توجهی در کیفیت تصاویر پُرونهایی دست‌یافته‌است. VITON-GT [11] با ارائه یک ماژول دفرمه‌سازی دو مرحله‌ای شامل تبدیل Affine و TPS، و یک ماژول پُرکردن هدایت‌شده با پارامترهای پیش‌بینی‌شده توسط تبدیل‌ها، نویزها را در تصاویر پُرونهایی کاهش می‌دهد. HR-VTON [12] برای بهبود کیفیت تصاویر پُرونهایی از ماژول اصلاح‌کننده (VDSR)[9] استفاده کرد. C-VTON [13] تنها بر قطعه‌بندی بدن شخص توسط DensePose تکیه داشت و از یک تولیدکننده تصویر قدرتمند با لایه‌های نرمال‌سازی شرطی بهره می‌برد. DP-VTON [14] نیز با معرفی یک ماژول همسان‌سازی لباس پس از ماژول دفرمه‌سازی، تلاش کرد تا جزئیات بیشتری از لباس هدف را حفظ کند.

روش‌های دو مرحله‌ای در پرو مجازی با چندین مشکل مواجه‌اند. یکی از اصلی‌ترین مشکلات این روش‌ها، دقت ناکافی در حفظ جزئیات بدن و لباس است. این روش‌ها اغلب به‌طور کامل نقشه‌های قطعه‌بندی بدن را را پوشش نمی‌دهند و برخی ویژگی‌های مهم لباس و بدن را از دست‌می‌دهند. افزون بر این، روش‌های دو مرحله‌ای معمولاً قادر نیستند لباس اولیه را به‌طور کامل حذف کنند که در نتیجه تصاویری با کیفیت پایین‌تر و کمتر واقع‌گرایانه تولید

---

[4] Alignment-Aware Segment Normalization
[5] Thin Plate Spline (TPS) Transformation
[6] Adversarial loss
[7] Grid Interval Consistency Loss (GIC)
[8] Occlusion Handling Technique (OHT)
[9] Very Deep Super Resolution

می‌کنند. این روش‌ها همچنین در مقابله با سبک‌های مختلف لباس و سایزهای متفاوت مشکل دارند، چرا که بیشتر بر شکل لباس تمرکز می‌کنند و به اطلاعات اندازه توجه کافی ندارند. به همین دلیل، تصاویر پرونهایی با دقت و کیفیت پایین‌تری تولید می‌شوند و این مسئله موجب کاهش تعمیم‌پذیری این روش‌ها به موقعیت‌های جدید می‌شود.

### 2-2- روش‌های سه‌مرحله‌ای
این روش‌ها شامل سه مرحله دفرمه‌سازی، پیش‌بینی نقشه قطعه‌بندی بدن و پروکردن می‌باشند. دو شاخه اصلی روش‌های سه مرحله‌ای به‌صورت زیر است.

**الف- روش‌هایی که ابتدا لباس هدف را دفرمه می‌کنند**
در این روش‌ها، ابتدا لباس هدف دفرمه می‌شود. سپس بر اساس لباس دفرمه‌شده، نقشه قطعه‌بندی بدن پیش‌بینی می‌شود و در نهایت، تصویر پرونهایی با استفاده از نقشه قطعه‌بندی پیش‌بینی‌شده و لباس دفرمه‌شده تولید می‌شود. به عنوان مثال، VTNFP [15] با استفاده از مکانیسم غیرمحلی[10] در ماژول دفرمه‌سازی، به بهبود فرآیند یادگیری و تطابق دقیق‌تر ویژگی‌ها می‌پردازد. این روش اولین مطالعه‌ای بود که با معرفی نقشه قطعه‌بندی بدن شخص در حالی که لباس هدف را برتن‌دارد، به‌عنوان راهنمایی برای تولید تصویر پرونهایی، کیفیت تصاویر پرونهایی را بهبود بخشید. روش LM-VTON [16] با معرفی تابع زیان مبتنی بر نقاط‌عطف، تغییرات ظریف‌تری را در اطراف لباس اعمال کرده و لباس‌های دفرمه را با مصنوعات کمتری تولید می‌کند. Dress Code [17] نیز با معرفی متمایزکننده معنایی‌آگاه‌در‌سطح‌پیکسل (PSAD)[11] امکان پرو هم‌زمان لباس‌های بالاتنه و پایین‌تنه را فراهم کرده است. این ویژگی به تولید نتایج واقعی‌تر و نزدیک‌تر به دنیای واقعی منجر شده است و به بهبود کیفیت تصاویر پرونهایی کمک کرده است.

**ب- روش‌هایی که ابتدا نقشه قطعه‌بندی بدن را پیش‌بینی می‌کنند**
در این روش‌ها، ابتدا نقشه قطعه‌بندی پیش‌بینی می‌شود. سپس بر اساس این نقشه قطعه‌بندی پیش‌بینی‌شده، لباس هدف دفرمه می‌شود و تصویر پرونهایی با استفاده از نقشه قطعه‌بندی پیش‌بینی‌شده و لباس دفرمه‌شده تولید می‌شود. برای مثال، ACGPN [18] برای تولید و حفظ محتوا در شبکه‌های پرو مجازی مبتنی بر تصاویر دو بعدی طراحی شده و یک محدودیت دیفرانسیل مرتبه دوم[12] برای کاهش مصنوعات در تصاویر دفرمه ارائه می‌دهد. VITON-HD [2] با استفاده از یک ژنراتور قدرتمند و لایه‌های نرمال‌سازی، موفق به تولید تصاویر پرو مجازی با وضوح ۷۶۸ × ۱۰۲۴ شده است. روش VITON-CROP [19] با استفاده از برش تصادفی تصاویر، امکان افزایش داده برای پرو مجازی با وضوح بالا را فراهم می‌کند. NL-VTON [20] با معرفی مکانیسم غیرمحلی و تابع زیان منظم‌سازی‌گریدی[13] در ماژول دفرمه‌سازی، لباس‌های دفرمه دقیق‌تری با حفظ بافت کلی و جزئیات محلی تولید می‌کند. AVTON [21] با معرفی ماژول پیش‌بینی اندام که تغییرات بین انواع لباس‌ها مانند آستین بلند به آستین کوتاه یا شلوار بلند به دامن کوتاه را درنظرمی‌گیرد و با یک ماژول دفرمه‌سازی بهبودیافته توسط تابع وندلند و یک بخش همجوشی[14] در ماژول پرونهایی به تصاویر واقعی‌تری دست یافته است. همچنین، WG-VITON [22] با افزودن یک ماسک باینری به ورودی‌ها، به بررسی مشکلات تنوع در سبک لباس پوشیدن در هنگام تعویض هم‌زمان لباس‌های بالاتنه و پایین‌تنه پرداخته است. برای مطالعه‌ای جامع‌تر و کلی درباره روش‌های پرو مجازی مبتنی بر تصویر، می‌توان به این مقاله مروری [23] نیز مراجعه کرد.

در نتیجه، مرور ادبیات نشان داد که پرو مجازی مبتنی بر تصاویر دو بعدی با بهره‌گیری از رویکردهای دومرحله‌ای و سه‌مرحله‌ای، منجر به بهبودهای قابل‌توجهی در کیفیت و دقت تصاویر پرونهایی شده است. این روش‌ها با پیش‌بینی نقشه‌قطعه‌بندی بدن و دفرمه‌سازی لباس‌ها، به تصاویر پرونهایی با دقت و واقع‌گرایی بیشتری دست یافته‌اند. همچنین، استفاده از تکنیک‌ها و الگوریتم‌های نوین، به ایجاد تصاویری با وضوح بالا و کیفیت بهتر کمک کرده است. این پیشرفت‌ها نه تنها تجربه کاربری را از پرو مجازی بهبود بخشیده، بلکه رضایت آن‌ها را نیز افزایش داده است. ادامه تحقیقات و توسعه تکنیک‌های جدید، افق‌های روشنی را برای آینده پرو مجازی نوید می‌دهد و انتظار می‌رود که این حوزه همچنان به رشد و پیشرفت‌های بیشتری دست‌یابد.

## 3- روش پیشنهادی
نمای کلی روش پیشنهادی در شکل ۲ نشان داده شده است. فرض کنید تصویر اولیه شخص هدف $I \in \mathbb{R}^{3 \times H \times W}$ و تصویر لباس هدف $c \in \mathbb{R}^{3 \times H \times W}$ را در اختیار داریم ($H$ و $W$ به ترتیب طول و عرض تصویر را نشان می‌دهند). هدف نهایی تولید تصویر مصنوعی $\hat{I} \in \mathbb{R}^{3 \times H \times W}$ است به‌طوری که در آن ژست و شکل بدن شخص هدف $I$ و همچنین ویژگی‌های لباس هدف $c$ حفظ شده است. آموزش مدل با سه‌تایی $(I, c, \hat{I})$ ساده است، اما ساخت

چنین مجموعه‌داده‌ای پرهزینه است. در عوض، از سه‌تایی $(I, c, I)$ که شخص هدف در تصویر اولیه قبلاً لباس هدف $c$ را پوشیده استفاده شده است. به دلیل اینکه آموزش مستقیم در سه‌تایی $(I, c, I)$ توانایی تعمیم مدل در زمان آزمایش را مختل می‌کند، نمایش مستقل از لباس شخص که در -VITON HD ارائه شده استفاده می‌شود. این نمایش اطلاعات لباس اولیه در تصویر شخص هدف $I$ را حذف کرده و به عنوان ورودی ماژول‌های بعدی استفاده می‌شود (بخش ۳-۱). در مرحله بعد، نقاط‌کلیدی لباس بر اساس ژست شخص هدف پیش‌بینی می‌شود و از این نقاط‌کلیدی پیش‌بینی‌شده به عنوان راهنمایی برای دفرمه‌سازی تصویر لباس هدف استفاده می‌شود (بخش ۳-۲). در ادامه لباس هدف مطابق نقاط‌کلیدی پیش‌بینی‌شده، دفرمه می‌شود (بخش ۳-۳). با توجه به لباس دفرمه‌شده بر اساس نقاط‌کلیدی پیش‌بینی‌شده و نمایش مستقل از لباس شخص، نقشه قطعه‌بندی بدن شخص در حالی که لباس هدف را برتن‌دارد، پیش‌بینی می‌شود (بخش ۳-۴) و در نهایت، با استفاده از نرمال‌سازی بخش‌های مرتب‌بشده اطلاعات گمراه‌کننده در نواحی نامرتب ایجادشده بین لباس دفرمه‌شده و ناحیه لباس پیش‌بینی‌شده توسط نقشه قطعه‌بندی پیش‌بینی‌شده حذف می‌شود. ژنراتور نواحی نامرتب را با بافت لباس هدف پر کرده و بدین ترتیب جزئیات لباس هدف را حفظ می‌کند (بخش ۳-۵).

### 3-1- ماژول نمایش مستقل از لباس شخص
نمایش مستقل از لباس شخص به معنای ارائه تصویری است که در آن لباس اولیه شخص هدف حذف شده و تنها اعضای بدن شخص به همراه ویژگی‌های ضروری آن حفظ شده است. به این ترتیب، مدل می‌تواند به‌درستی بر روی لباس هدف تمرکز و بدون تأثیر از لباس‌های اولیه، لباس‌های هدف را جایگزین کند. برای آموزش مدل با لباس هدف $c$ و تصویر شخص هدف $I$ که لباس هدف $c$ را قبلاً پوشیده است، از یک نمایش مستقل از لباس شخص در مدل‌های پرو مجازی استفاده شده است. چنین نمایش شخصی باید شرایط زیر را داشته باشد:

- لباس اولیه شخص باید کاملاً حذف شود و اثراتی از لباس اولیه باقی نماند. این اقدام، مدل را قادر می‌سازد تا بدون تأثیر از لباس اولیه، بر روی جایگزینی دقیق لباس هدف تمرکز کند.
- اطلاعات موردنیاز برای پیش‌بینی ژست و شکل بدن شخص باید حفظ شود. این شامل ویژگی‌هایی مانند قوس‌های بدن، اعضای بدن و وضعیت ایستادن شخص است. همچنین این امر امکان پیش‌بینی صحیح چگونگی قرارگیری لباس هدف بر روی بدن شخص را فراهم می‌کند.
- حفظ مناطقی مانند سر، صورت و موها برای شناسایی هویت شخص هدف ضروری است. این امر مدل را قادر می‌سازد تا هویت شخص را شناسایی و تصویر پرونهایی واقعی‌تری تولید کند.
- برای پرو لباس‌های بالاتنه به تنهایی، لازم است که لباس‌های پایین‌تنه در نتایج پرونهایی بدون تغییر باقی بمانند. این تضمین می‌کند که تغییرات فقط در قسمت بالاتنه اعمال شوند و سایر اعضای بدن شخص هدف بدون تغییر باقی بمانند.

در این نمایش مستقل از لباس شخص، یک تصویر مستقل از لباس شخص $I_a$ و یک نقشه قطعه‌بندی مستقل از لباس شخص $S_a$ به عنوان ورودی برای ماژول‌های بعدی پیشنهاد شده است، که در آن لباس اولیه شخص هدف حذف شده و اعضای بدن شخص که نیاز به بازتولید دارند، حفظ می‌شوند. ابتدا نقشه قطعه‌بندی $S \in \mathbb{L}^{H \times W}$ و نقشه ژست $P \in \mathbb{R}^{3 \times H \times W}$ تصویر شخص هدف $I$ با استفاده از شبکه‌های ازپیش‌آموزش‌دیده [24] و [25] پیش‌بینی می‌شوند ($\mathbb{L}$ مجموعه‌ای از اعداد صحیح است که برچسب‌های معنایی را نشان می‌دهد). نقشه قطعه‌بندی مستقل از لباس شخص $S_a$ برای حذف ناحیه لباس اولیه و حفظ بقیه اعضای بدن شخص استفاده می‌شود. به این ترتیب، مدل تنها بر روی بدن شخص تمرکز می‌کند و اطلاعات اضافی حذف می‌شود. نقشه ژست $P$ برای برداشتن بازوها استفاده می‌شود، اما نه دست‌ها، زیرا بازتولید دست‌ها به‌طور دقیق در نتایج پرونهایی دشوار است. برای حذف وابستگی به لباسی که در ابتدا توسط شخص هدف پوشیده شده است، مناطقی که می‌توانند هرگونه اطلاعات لباس اولیه را ارائه دهند (مانند بازوهایی که به طول آستین اشاره می‌کنند)، باید حذف شوند. بنابراین، هنگام ایجاد یک تصویر مستقل از لباس شخص $I_a$، بازوها از تصویر شخص موردنظر $I$ حذف می‌شوند. مناطق مربوط به لباس اولیه و بازوها با رنگ خاکستری ماسک(پوشانده) می‌شوند، به طوری که پیکسل‌های ماسک‌شده تصویر دارای مقدار صفر هستند. به ماسک‌ها padding (پیکسل‌های اضافی در لبه‌ها) اضافه می‌شود تا اطمینان حاصل شود نواحی لباس اولیه کاملاً از بین بروند. در اینجا عرض padding به صورت تجربی تعیین می‌شود. استفاده از نمایش مستقل از لباس شخص به مدل اجازه می‌دهد تا با دقت بیشتری

---
[10] Non-local Mechanism (NL)
[11] Pixel-wise Semantic-Aware Discriminator
[12] Second-order Difference Constraint
[13] Grid Regularization Loss Function
[14] Fusion Part

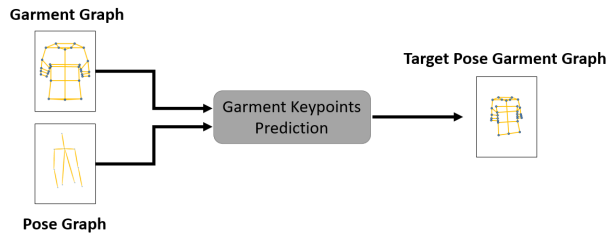

۲- ماژول پیش‌بینی نقاط‌کلیدی لباس

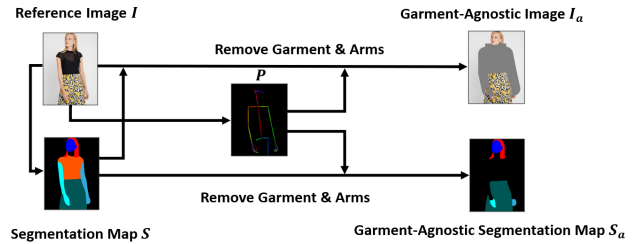

۱- ماژول نمایش مستقل از لباس شخص

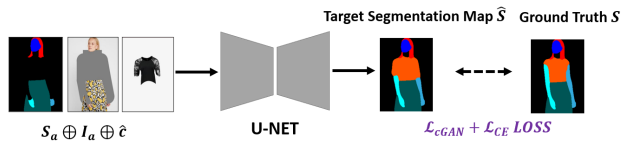

۴- ماژول پیش‌بینی نقشه قطعه‌بندی بدن شخص هدف

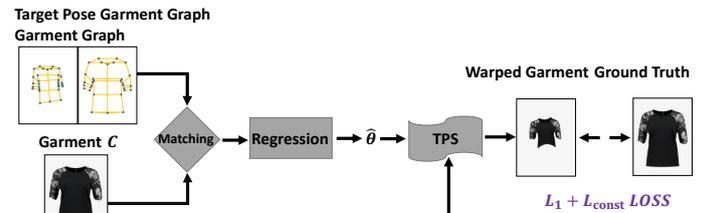

۳- ماژول دفرمه‌سازی تصویر لباس هدف

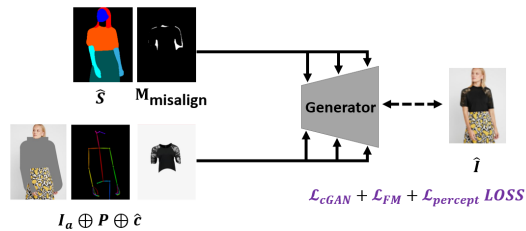

۵- ماژول پرو کردن

شکل ۲- نمای کلی روش +VITON-HD پیشنهادی که از پنج ماژول تشکیل شده است.

لباس‌های هدف را بر روی بدن اشخاص جایگزین کند. با حذف کامل لباس اولیه و حفظ اطلاعات مهم بدن و هویت شخص، مدل به نتایج واقع‌گرایانه‌تری دست می‌یابد.

**۳-۲- ماژول پیش‌بینی نقاط‌کلیدی لباس**

برای پیش‌بینی نقاط‌کلیدی لباس، این مسئله به عنوان یک رگرسیون گراف لباس وابسته به گراف ژست فرموله می‌شود. برای حل این مسئله، از یک شبکه عصبی گراف دوجریانی ارائه‌شده در [26] استفاده شده است. این شبکه برای پردازش اطلاعات گراف‌های لباس و ژست طراحی‌شده و شامل بلوک‌های کانولوشنی گراف است که با اعمال فیلتر بر گره‌ها و یال‌های گراف، ویژگی‌های محلی را استخراج می‌کند. نقاط‌کلیدی با استفاده از روش پیشرفته معرفی‌شده در [27] استخراج می‌شوند. شبکه عصبی گراف دوجریانی، اطلاعات گراف‌های لباس و ژست را به طور همزمان پردازش می‌کند و از بلوک‌های کانولوشنی گراف برای شناسایی الگوهای پیچیده و بهبود دقت پیش‌بینی نقاط‌کلیدی لباس استفاده می‌کند. این روش به مدل این امکان را می‌دهد که با دقت بیشتری نقاط‌کلیدی لباس را پیش‌بینی کرده و لذا عملکرد بهتری در دفرمه‌سازی لباس‌ها هدف داشته باشد.

**۳-۳- ماژول دفرمه‌سازی تصویر لباس هدف**

طراحی ماژول دفرمه‌سازی پیشنهادی از ساختار ماژول دفرمه‌سازی در CP-VTON الهام گرفته شده است، اما با رویکردی نوآورانه که به‌جای استفاده از لایه‌های استخراج ویژگی و محاسبه ماتریس همبستگی میان ویژگی‌ها، تنها از نقاط کلیدی بهره می‌گیرد. این ماژول، لباس هدف $c$ را با استفاده از نقاط‌کلیدی پیش‌بینی‌شده توسط ماژول قبلی دفرمه می‌کند. در این روش‌پیشنهادی، نقاط کلیدی که موقعیت‌های مهم لباس (مانند یقه، آستین‌ها و لبه‌های پایین لباس) را مشخص می‌کنند، مستقیماً به ماژول دفرمه‌سازی منتقل می‌شوند. بهره‌گیری از این اطلاعات دقیق، امکان اعمال تغییرات هندسی مورد نیاز با سطح بالایی از دقت را فراهم می‌سازد. حذف نمایش مستقل از لباس شخص و جایگزینی آن با نقاط‌کلیدی پیش‌بینی‌شده، وابستگی مدل به لباس اولیه را به حداقل رسانده است. این تغییر همچنین منجر به حذف لایه‌های غیرضروری استخراج ویژگی شده و در نهایت، باعث کاهش پیچیدگی مدل

گردیده است. ورودی‌های ماژول دفرمه‌سازی شامل لباس هدف $c$، نقاط‌کلیدی پیش‌بینی‌شده لباس و نقاط‌کلیدی لباس هدف است. در گام نخست، یک ماتریس انطباق مستقیم میان هر سه ورودی محاسبه می‌شود تا ارتباط و همبستگی هندسی و مکانی میان این اجزا به دقت مدل‌سازی شود. این ماتریس سپس به شبکه رگرسیون ارسال می‌گردد تا پارامترهای تبدیل TPS، $\theta \in \mathbb{R}^{2 \times 5 \times 5}$ را پیش‌بینی کند. پارامترهای پیش‌بینی‌شده برای دفرمه‌سازی لباس هدف $c$ استفاده می‌شوند و باعث انطباق لباس به شکلی متناسب با ژست و شکل بدن شخص هدف می‌شوند. ماژول دفرمه‌سازی پیشنهادی توسط تابع زیان نرم بین یک لباس‌های دفرمه‌شده $\hat{c}$ و لباس‌های ایده آل $I_c$ که از تصویر شخص هدف $I$ استخراج شده، آموزش داده می‌شود. علاوه‌براین، یک محدودیت دیفرانسیل مرتبه دوم ارائه‌شده در [18] برای کاهش مصنوعات در تصاویر لباس‌های دفرمه‌شده اتخاذ شده است. این محدودیت به حفظ یکپارچگی و همواری سطوح لباس هنگام اعمال دفرمه‌سازی کمک می‌کند و از تغییرات ناگهانی و غیرطبیعی جلوگیری می‌کند. تابع زیان نهایی $\mathcal{L}_{warp}$ برای دفرمه‌سازی لباس هدف متناسب با ژست و شکل بدن شخص هدف در رابطه (۱) نشان داده شده است.

$$\mathcal{L}_{warp} = \|I_c - \hat{c}\|_1 + \lambda_{const} \mathcal{L}_{const} \qquad (1)$$

که در رابطه (۱)، $\hat{c}$ لباس دفرمه‌شده، $\mathcal{L}_{const}$ یک محدودیت دیفرانسیل مرتبه دوم و $\lambda_{const}$ به عنوان ابرپارامتر برای $\mathcal{L}_{const}$ در نظر گرفته شده است.

**۳-۴- ماژول پیش‌بینی نقشه قطعه‌بندی بدن شخص هدف**

با توجه به نقشه قطعه‌بندی مستقل از لباس شخص و نقشه ژست به هم ترکیب شده $(S_a, P)$ و لباس دفرمه‌شده $\hat{c}$، ژنراتور قطعه‌بندی $G_S$ نقشه قطعه‌بندی بدن شخص $\hat{S} \in \mathbb{L}^{H \times W}$ در حالی که لباس هدف $c$ بر تن دارد، پیش‌بینی می‌کند. ژنراتور قطعه‌بندی از معماری U-Net [5] بهره می‌برد، که شامل لایه‌های کانولوشن، لایه‌های Downsampling و لایه‌های Upsampling می‌باشد. دو متمایزکننده چندمقیاسی برای تابع زیان خصمانه مشروط به کار گرفته شده است. این متمایزکننده‌ها به مدل کمک می‌کنند تا جزئیات و

ویژگی‌های مختلف در مقیاس‌های مختلف را بهبود بخشد. برای مطالعه جزئیات بیشتر در مورد متمایزکننده به مقاله [28] مراجعه شود. تابع زیان نهایی $\mathcal{L}_{Seg}$ در ماژول پیش‌بینی نقشه قطعه‌بندی بدن شخص هدف در رابطه (۲) نشان داده شده است.

$$\mathcal{L}_{Seg} = \mathcal{L}_{cGAN} + \lambda_{CE}\mathcal{L}_{CE} \tag{۲}$$

در رابطه (۲) $\mathcal{L}_{CE}$ و $\mathcal{L}_{cGAN}$ به ترتیب تابع زیان آنتروپی متقاطع پیکسل به پیکسل و تابع زیان خصمانه مشروط بین نقشه قطعه‌بندی پیش‌بینی‌شده $\hat{S}$ و نقشه قطعه‌بندی بدن شخص هدف $S$ را نشان می‌دهند. تابع زیان $\mathcal{L}_{CE}$ با هدف تولید نقشه قطعه‌بندی دقیق‌تر، اختلاف بین احتمال پیش‌بینی‌شده و احتمال واقعی را به‌ازای هر پیکسل در نقشه قطعه‌بندی محاسبه می‌کند و سپس میانگین این اختلاف‌ها را به عنوان زیان نهایی ارائه می‌دهد. $\mathcal{L}_{cGAN}$ از نوع [29] LSGAN است، که هدف آن تولید نقشه‌های قطعه‌بندی بدن با شباهت بالا به نقشه‌های واقعی است که به جای استفاده از آنتروپی متقاطع معمول از خطای میانگین مربعات خطاها برای بهبود پایداری و کیفیت نتایج پُرونهایی استفاده می‌کند. $\lambda_{CE}$ ابرپارامتر مربوط به تابع زیان آنتروپی متقاطع پیکسل به پیکسل است، که به تعادل این توابع زیان و بهبود عملکرد کلی مدل کمک می‌کند. روابط ریاضی توابع زیان $\mathcal{L}_{CE}$ و $\mathcal{L}_{cGAN}$ به‌صورت زیر ارائه شده است.

$$\mathcal{L}_{CE} = -\frac{1}{HW}\sum_{k\in C, y\in H, x\in W} S_{k,y,x}\log(\hat{S}_{k,y,x}) \tag{۳}$$

$$\mathcal{L}_{cGAN} = \mathbb{E}_{(X,S)}[\log(D(X,S))] + \mathbb{E}_X[1 - \log(D(X,\hat{S}))] \tag{۴}$$

در رابطه (۳)، $S_{k,y,x}$ و $\hat{S}_{k,y,x}$ مقادیر پیکسل نقشه قطعه‌بندی بدن شخص هدف $S$ و نقشه قطعه‌بندی پیش‌بینی‌شده $\hat{S}$ مربوط به مختصات $(x,y)$ در کانال $k$ را نشان می‌دهند. نمادهای $H$، $W$ و $C$ به ترتیب نشان‌دهنده ارتفاع، عرض و تعداد کانال‌های نقشه قطعه‌بندی $S$ هستند. در رابطه (۴)، $X$ ورودی‌های ژنراتور قطعه‌بندی ($S_a, P, \hat{c}$) را نشان می‌دهد و نماد $D$ نشان‌دهنده متمایزکننده است. همچنین، $\mathbb{E}_X$ امیدریاضی بر روی داده های $X$ و $\mathbb{E}_{(X,S)}$ امیدریاضی بر روی جفت‌های $(X,S)$ را نشان می‌دهد.

### ۳-۵- ماژول پُروکردن

ماژول پُروکردن در این روش‌پیشنهادی شامل دو مرحله است: نرمال‌سازی بخش‌های مرتب‌شده و ژنراتور که در ادامه این دو بخش را تشریح می‌پردازیم.

**الف- نرمال‌سازی بخش‌های مرتب‌شده**

برای بهبود دقت و کیفیت تصاویر پُرونهایی، از تکنیک پیشرفته نرمال‌سازی بخش‌های مرتب‌شده که در VITON-HD ارائه‌شده، استفاده می‌شود. این تکنیک نقشه قطعه‌بندی پیش‌بینی‌شده $\hat{S}$ را به صورت دقیق و واقعی‌گرایانه تنظیم می‌کند تا از انتقال نادرست اطلاعات جلوگیری شود و نواحی مختلف بدن و لباس به درستی شناسایی و پردازش شوند. فرآیند نرمال‌سازی بخش‌های مرتب‌شده دو ورودی دارد: ۱- نقشه قطعه‌بندی پیش‌بینی‌شده و ۲- ماسک باینری نواحی نامرتب $M_\text{misalign}$، که با حذف ماسک لباس دفرمه از ناحیه لباس هدف $\hat{S}_c$ در نقشه قطعه‌بندی پیش‌بینی‌شده، به دست می‌آید. رابطه (۵) نحوه محاسبه $M_\text{align}$ و $M_\text{misalign}$ را نمایش می‌دهد ($M_{\hat{c}}$ نشان‌دهنده ماسک لباس دفرمه است).

$$\begin{aligned}M_\text{align} &= \hat{S}_c \cap M_{\hat{c}}\\ M_\text{misalign} &= \hat{S}_c - M_\text{align}\end{aligned} \tag{۵}$$

نسخه اصلاح شده $\hat{S}$ به صورت $\hat{S}_{div}$ تعریف شده است، که در آن ناحیه لباس هدف $\hat{S}_c$ در نقشه قطعه‌بندی پیش‌بینی‌شده $\hat{S}$ به $M_\text{align}$ (نواحی هماهنگ و مرتب) و $M_\text{misalign}$ (نواحی نامرتب) تقسیم می شود. در این مرحله نواحی نامرتب $M_\text{misalign}$ و سایر نواحی در $h^i$ به‌طور جداگانه نرمال‌سازی می شوند و سپس ویژگی‌های تنظیم شده با استفاده از پارامترهای تبدیل affine که از $\hat{S}_{div}$ استخراج شده‌اند، تعدیل می‌شوند. در شکل ۳، $h^i \in \mathbb{R}^{N\times C^i \times H^i \times W^i}$ به عنوان پاسخ لایه i-اُم شبکه برای یک دسته از نمونه‌های $N$ نشان داده شده است،

جایی که $H^i$، $W^i$ و $C^i$ به ترتیب ارتفاع، عرض و تعداد کانال‌های $h^i$ را نشان می‌دهند. $N$ تعداد نمونه‌ها در یک دسته (بچ) است که به معنای پردازش همزمان شبکه بر روی $N$ نمونه می‌باشد. مقدار پاسخ در موقعیت $(n \in N, k \in C^i, y \in H^i, x \in W^i)$ توسط رابطه (۶) محاسبه می‌شود.

$$\gamma^i_{k,y,x}(\hat{S}_{div})\frac{h^i_{n,k,y,x} - \mu^{i,m}_{n,k}}{\sigma^{i,m}_{n,k}} + \beta^i_{k,y,x}(\hat{S}_{div}) \tag{۶}$$

در رابطه (۶) $h^i_{n,k,y,x}$ پاسخ در موقعیت $(n,k,y,x)$ قبل از نرمال‌سازی است. این مقادیر به عنوان ورودی به فرآیند نرمال‌سازی وارد می‌شوند. فرآیند نرمال‌سازی به معنای تعدیل پاسخ‌ها با استفاده از میانگین و انحراف معیار آنهاست، به گونه‌ای که داده‌ها به یک مقیاس مشترک تبدیل شوند. $\gamma^i_{k,y,x}$ و $\beta^i_{k,y,x}$ توابعی هستند که نقشه قطعه‌بندی $\hat{S}_{div}$ را به پارامترهای مدولاسیون لایه نرمال‌سازی تبدیل می‌کنند. به‌عبارتی، این پارامترها به تنظیم دقیق فعال‌سازی‌های نرمال‌شده کمک می‌کنند. $\mu^{i,m}_{n,k}$ و $\sigma^{i,m}_{n,k}$ به ترتیب میانگین و انحراف معیار پاسخ در نمونه $n$ و کانال $k$ هستند. میانگین نمایانگر مقدار متوسط پاسخ‌ها و انحراف‌معیار نمایانگر پراکندگی پاسخ‌هاست. $\mu^{i,m}_{n,k}$ و $\sigma^{i,m}_{n,k}$ به ترتیب توسط روابط (۷) و (۸) محاسبه می‌شوند، که این روابط به‌طور دقیق فرآیند محاسبه این پارامترها را توضیح می‌دهند.

$$\mu^{i,m}_{n,k} = \frac{1}{|\Omega^{i,m}_n|}\sum_{(y,x)\in\Omega^{i,m}_n} h^i_{n,k,y,x} \tag{۷}$$

$$\sigma^{i,m}_{n,k} = \sqrt{\frac{1}{|\Omega^{i,m}_n|}\sum_{(y,x)\in\Omega^{i,m}_n}\left(h^i_{n,k,y,x} - \mu^{i,m}_{n,k}\right)^2} \tag{۸}$$

$\Omega^{i,m}_n$ مجموعه پیکسل ها در ناحیه $m$ را نشان می دهد، که ممکن است $M_\text{misalign}$ یا ناحیه دیگری باشد و $|\Omega^{i,m}_n|$ تعداد پیکسل ها در $\Omega^{i,m}_n$ است. نرمال‌سازی بخش‌های مرتب‌شده با تفکیک و نرمال‌سازی جداگانه نواحی نامرتب و سایر نواحی به بهبود دقت و کیفیت تصاویر پُرونهایی کمک می‌کند. این فرآیند با تضمین حذف اطلاعات گمراه‌کننده، تصاویر پُرونهایی واقع‌گرایانه‌تر و با کیفیت بالاتری تولید می‌کند.

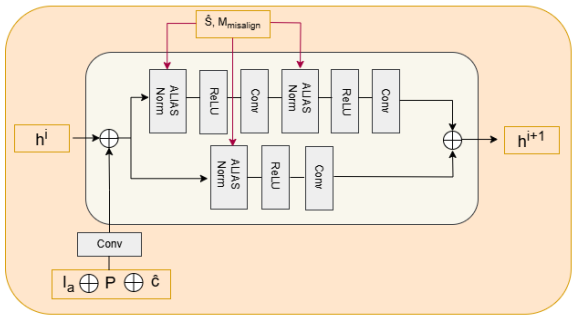

**شکل ۳- نمای جزئی از یک بلوک ResBlk. ورودی تغییر اندازه داده شده ($I_a, P, \hat{c}$) پس از عبور از یک لایه کانولوشن به $h^i$ الحاق می‌شود. هر لایه نرمال‌سازی بخش‌های مرتب‌شده از $\hat{S}$ و $M_\text{misalign}$ تغییر اندازه داده شده برای نرمال‌سازی پاسخ‌ها استفاده می‌کند.**

**ب- ژنراتور**

در نهایت، هدف تولید تصویر مصنوعی نهایی $\hat{I}$ بر اساس خروجی‌های ماژول‌های قبل است. این تصویر نهایی باید ژست و شکل بدن شخص را حفظ کرده و ویژگی‌های لباس هدف را به‌درستی نمایش دهد. به‌طور کلی، نمایش مستقل از لباس شخص $I_a$، نقشه ژست $P$ و تصویر لباس دفرمه‌شده $\hat{c}$ با هم ترکیب می‌شوند. مقدار $(I_a, P, \hat{c})$ ترکیب شده به هر لایه ژنراتور تزریق می‌شود، که توسط $\hat{S}$ هدایت می‌کند تا اطلاعات مربوط به بدن شخص، ژست و لباس دفرمه‌شده را به‌طور همزمان پردازش کند. شکل ۴ نمای کلی از ژنراتور را شرح می‌دهد، که در آن معماری ساده‌شده‌ای اتخاذ شده است که بخش رمزگذار در ساختار شبکه رمزگذار-رمزگشا حذف شده و تنها

بخش رمزگشا برای پردازش مورد استفاده قرار می‌گیرد. این روش باعث کاهش پیچیدگی مدل و بهبود کارایی آن می‌شود.

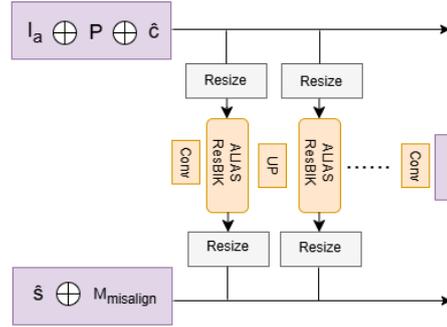

**شکل 4- نمای کلی ژنراتور - این ژنراتور شامل مجموعه‌ای از بلوک‌های ResBlk و لایه‌های upsampling می‌باشد. ورودی($I_a, P, \hat{c}$) تغییر اندازه داده می‌شود و به هر لایه ژنراتور تزریق می‌شود.**

ژنراتور از یک سری بلوک‌های ResBlk (residual block) با لایه‌های upsampling استفاده می‌کند. هر بلوک ResBlk، همان‌طور که در شکل 3 نشان داده شده است، از سه لایه کانولوشنال و سه لایه نرمال‌سازی تشکیل شده است. بلوک‌های ResBlk در سطوح مختلف رزولوشن فعالیت می‌کنند. هر بلوک برای تصاویر با اندازه‌های متفاوت، نیاز به تغییر اندازه ورودی دارد تا اطلاعات به‌درستی مقیاس‌بندی شود و بهترین کارایی و دقت در پردازش ارائه گردد. با توجه به رزولوشن‌های متفاوتی که بلوک‌های ResBlk در آن‌ها فعالیت دارند، اندازه ورودی لایه‌های نرمال‌سازی، $\hat{S}$ و $M_{\text{misalign}}$ قبل از تزریق به هر لایه تغییر داده می‌شود. این کار تضمین می‌کند که اطلاعات ورودی به‌درستی پردازش شوند و کیفیت نهایی تصویر حفظ شود. به‌طور مشابه، ورودی ژنراتور شامل ($I_a, P, \hat{c}$)، ابتدا متناسب با نیازهای رزولوشنی بلوک‌های ResBlk تغییر اندازه داده و تنظیم می‌شود. این ورودی‌های پردازش‌شده، پس از عبور از یک لایه کانولوشن برای استخراج ویژگی‌های مهم، با پاسخ لایه‌های قبلی ترکیب می‌گردند. این ترکیب به هر بلوک ResBlk امکان می‌دهد تا با استفاده از اطلاعات جدید و داده‌های پیشین، خروجی خود را با دقت و کارایی بیشتری تنظیم و تولید کند. شبکه با استفاده از تکنیک اصلاح چند مقیاسی در سطح ویژگی، جزئیات لباس را دقیق‌تر از اصلاح در سطح پیکسل حفظ می‌کند. این اصلاح، پردازش اطلاعات را یکپارچه‌تر انجام می‌دهد و مدل با درک بهتر روابط بین ویژگی‌های مختلف، از گم شدن جزئیات مهم جلوگیری می‌کند. در نتیجه، وضوح و کیفیت تصاویر پرونهایی بهبود می‌یابد. تابع زیان ژنراتور از الگوریتم SPADE [30] و pix2pixHD [28] بهره می‌برد تا تصاویری واقع‌گرایانه با وضوح بالا و جزئیات غنی تولید کند. این فرآیند شامل تابع زیان خصمانه مشروط $\mathcal{L}_{cGAN}$، تابع زیان تطبیق ویژگی $\mathcal{L}_{FM}$ و تابع زیان ادراکی $\mathcal{L}_{\text{percept}}$ است که هر کدام نقش منحصر به فردی در بهبود کیفیت و هماهنگی تصاویر نهایی دارند. تابع زیان نهایی $\mathcal{L}_{TON}$ برای ژنراتور در رابطه (9) نشان داده شده است.

$$\mathcal{L}_{TON} = \mathcal{L}_{cGAN} + \lambda_{FM}\mathcal{L}_{FM} + \lambda_{\text{percept}}\mathcal{L}_{\text{percept}}, \quad (9)$$

که در رابطه (9)، $\lambda_{FM}$ و $\lambda_{\text{percept}}$ ابرپارامترها هستند. در ادامه به چگونگی محاسبه سه تابع زیان که در رابطه (9) ذکر شده‌اند، می‌پردازیم:

- **تابع زیان $\mathcal{L}_{cGAN}$** که از رابطه (10) محاسبه می‌شود، به واقعی‌تر شدن تصاویر کمک می‌کند. در این رابطه $I$ تصویر شخص هدف، $c$ تصویر لباس هدف، $\hat{I}$ تصویر نهایی تولیدشده توسط ژنراتور و $D_I$ متمایزکننده می‌باشد. $S_{div}$ نسخه اصلاح‌شده نقشه قطعه‌بندی $S$ است.

$$\mathcal{L}_{cGAN} = \mathbb{E}_I\big[\log\big(D_I(S_{\text{div}}, I)\big)\big] + \mathbb{E}_{(I,c)}\big[1 - \log\big(D_I(S_{\text{div}}, \hat{I})\big)\big] \quad (10)$$

- **تابع زیان $\mathcal{L}_{FM}$** باعث افزایش شباهت ویژگی‌های تصویر تولیدشده به تصویر واقعی می‌شود. این تابع زیان در رابطه (11) آورده شده است که در آن $T$ تعداد لایه‌های استفاده شده در متمایزکننده $D_I$ است و $D_I^{(i)}$ و $K_i$ به ترتیب پاسخ و تعداد عناصر در لایه $i$-اُم در $D_I$ هستند.

$$\mathcal{L}_{FM} = \mathbb{E}_{(I,c)}\sum_{i=1}^{T}\frac{1}{K_i}\big[\|D_I^{(i)}(S_{\text{div}}, I) - D_I^{(i)}(S_{\text{div}}, \hat{I})\|_1\big] \quad (11)$$

- **تابع زیان $\mathcal{L}_{\text{percept}}$** که تابع زیان ادراکی است به بهبود جزئیات و کیفیت بصری تصاویر نهایی تمرکز دارد. رابطه (12) این تابع زیان را بیان می‌کند که در این رابطه $V$ تعداد لایه‌های استفاده شده در شبکه VGG، $F$ است (برای جزئیات بیشتر به مقاله [31] مراجعه شود) و $F^{(i)}$ و $R_i$ به‌ترتیب پاسخ و تعداد عناصر موجود در لایه $i$-اُم در $F$ هستند.

$$\mathcal{L}_{\text{percept}} = \mathbb{E}_{(I,c)}\sum_{i=1}^{V}\frac{1}{R_i}\big[\|F^{(i)}(I) - F^{(i)}(\hat{I})\|_1\big] \quad (12)$$

در رابطه (12) تابع زیان خصمانه استاندارد با Hinge loss [32] جایگزین شده است. این جایگزینی باعث می‌شود که مدل پایداری بیشتری داشته باشد و نوسانات در طی فرآیند آموزش کاهش یابد، که به بهبود کیفیت و دقت تصاویر تولیدشده کمک می‌کند. درنهایت، روش VITON-HD+ پیشنهادی قادر است با دقت بیشتری ویژگی‌های لباس هدف را حفظ کند و تصاویری با کیفیت بالا تولید کند. این روش‌ پیشنهادی چشم‌اندازی امیدوارکننده برای بهبود بیشتر در زمینه پرو مجازی مبتنی بر تصویر و تولید تصاویر واقع‌گرایانه ارائه می‌دهد.

## 4- آزمایشات

### 4-1- مجموعه داده

در این مقاله، از مجموعه‌داده VITON-HD با رزولوشن 1024 × 768 استفاده شده است. این مجموعه‌داده شامل 13,679 جفت تصویر از نما جلو زنان و لباس‌های بالاتنه آن‌هاست که از یک وب‌سایت خرید آنلاین گردآوری شده است. این جفت‌ها به دو دسته تقسیم شدند: 11,647 جفت برای آموزش و 2,032 جفت برای آزمایش. برای ارزیابی حالت جفت‌شده، از جفت‌های تصویر شخص و لباس استفاده شده است. در حالت غیرجفت‌شده، تصاویر لباس‌ها به صورت تصادفی با تصاویر اشخاص ترکیب شدند. در حالت جفت‌شده، تصویر شخص با لباس اصلی بازسازی می‌شود، در حالی که در حالت غیرجفت‌شده، لباس تصویر شخص با یک لباس متفاوت جایگزین می‌شود.

### 4-2- مقایسه کیفی با روش‌های پیشرفته حال حاضر

هدف این بخش مقایسه عملکرد و کیفیت روش‌پیشنهادی با روش‌های مختلف در تولید تصاویر پُرو مجازی است. روش پیشنهادی با دو روش CP-VTON و VITON-HD مورد مقایسه قرار گرفته است. CP-VTON به عنوان یک روش پایه، دو ماژول دفرمه‌سازی و پُروکردن را معرفی کرده است و تمامی روش‌های موجود مبتنی بر تبدیل TPS همین دو مرحله را به‌عنوان مبنا قرار داده‌اند و با مکانیسم‌های مختلف بهبود بخشیده‌اند. VITON-HD نیز به عنوان یک روش پیشرفته شناخته می‌شود، که یک استراتژی سه‌مرحله‌ای شامل پیش‌بینی نقشه قطعه‌بندی بدن، دفرمه‌سازی و پُروکردن لباس هدف را دنبال می‌کند. تمامی نتایج آزمایش‌های CP-VTON و VITON-HD از طریق دانلود و اجرای کدهای منبع آن‌ها به دست آمده‌اند.

با تحلیل دقیق نتایج، نشان می‌دهیم که روش VITON-HD+ چگونه در حفظ ویژگی‌های لباس هدف، از جمله شکل و بافت، و همچنین تعمیم‌پذیری به موقعیت‌های جدید، عملکرد بهتری دارد. مطابق با شکل 5 ردیف (1)، در شرایطی که شخص ابتدا لباس آستین‌کوتاهی پوشیده و لباس آستین‌بلندی با طرح‌های تکراری در بافت لباس را برای پُرو انتخاب می‌کند، مشاهده می‌شود که تصویر پُرونهایی تولیدشده توسط CP-VTON از واقعیت فاصله زیادی دارد و این روش قادر به پُرو صحیح لباس هدف نبوده و درنتیجه شکل و جزئیات لباس هدف را از دست می‌دهد. در مقابل، در تصویر نهایی تولیدشده توسط VITON-HD، هرچند شکل کلی لباس حفظ شده، اما جزئیات لباس هدف به‌طور کامل نشان داده نشده‌اند. در مقابل، روش VITON-HD+ پیشنهادی نه تنها لباس هدف را به‌درستی پُرو کرده، بلکه جزئیات بیشتری از لباس هدف را نیز حفظ کرده است. در ردیف (2)، لباس خردلی رنگ با بافت ریز و طرح‌دار توسط VITON-HD+ پیشنهادی با وضوح بالا تولید شده و جزئیات اطراف یقه به خوبی نمایش داده شده‌اند، در حالی که روش VITON-HD نتوانسته این جزئیات را خیلی واضح نمایش دهد. همچنین، روش CP-VTON در مواجهه با لباس‌هایی که دارای جزئیات پیچیده هستند، نتایج موفقیت‌آمیزی ندارد. در ردیف (3)، آرم سفید رنگ لباس هدف با استفاده از روش VITON-HD+ به‌طور قابل‌توجهی بهتر از روش VITON-HD حفظ شده است. در این مورد نیز، تصویر نهایی تولیدشده توسط CP-VTON کیفیت لازم را ندارد. در ردیف (4)، لوگو چاپ شده بر لباس هدف با استفاده از روش VITON-HD+ پیشنهادی در مقایسه با VITON-HD در ابعاد واقعی‌تری تولید شده است. در مقابل، در روش CP-VTON تصویر پُرونهایی تار و بی‌کیفیت است همچنین پشت یقه در تصویر پُرونهایی دیده می‌شود. همان‌طور که در ردیف (5) نشان داده شده است، لباس هدف دارای یقه هفت می‌باشد.

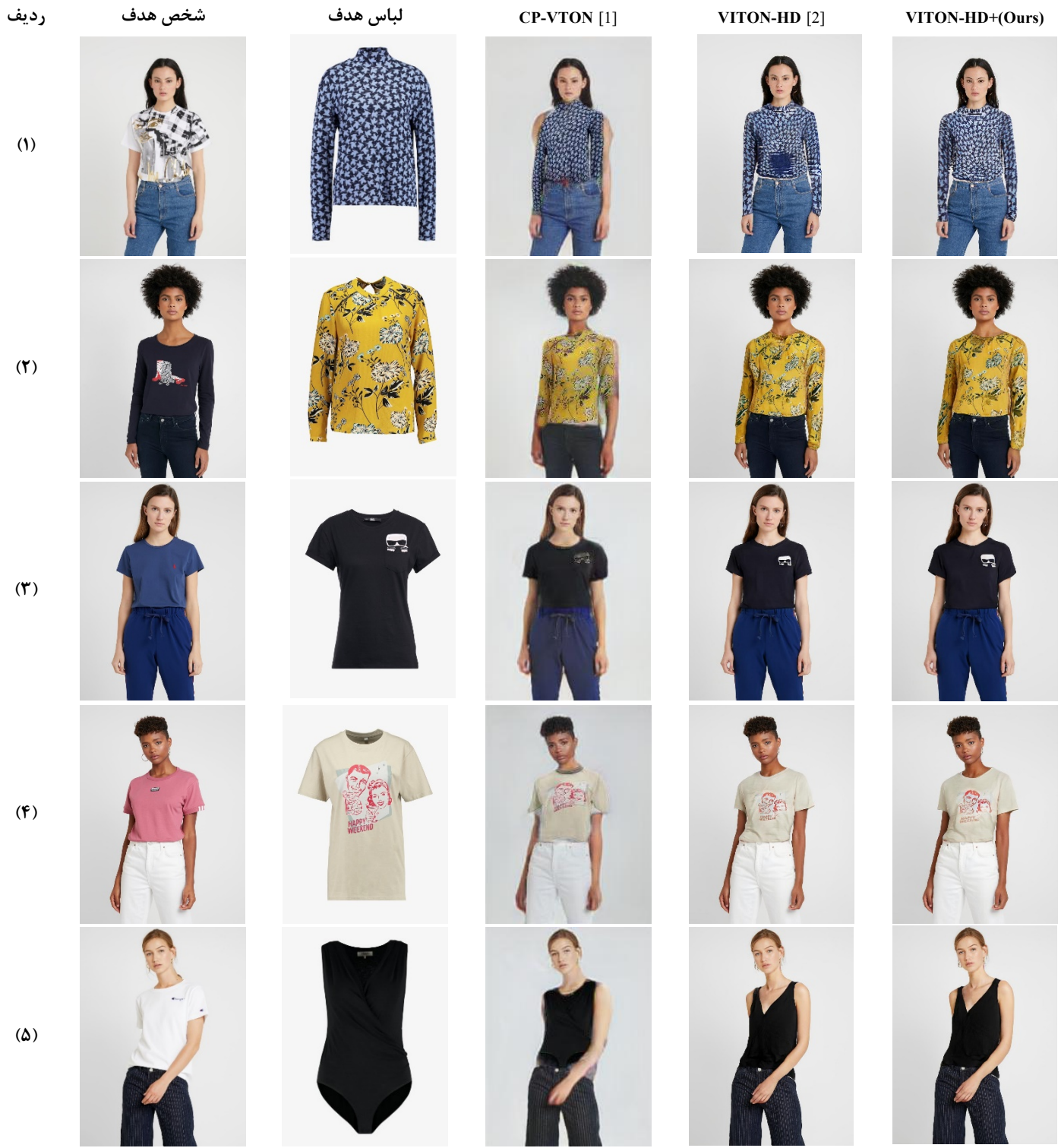

شکل۵- مقایسه کیفی روش های CP-VTON، VITON-HD و +VITON-HD پیشنهادی.

تصویر نهایی تولیدشده توسط CP-VTON در نواحی یقه به لباس اولیه وابسته است، در حالی که روش VITON-HD توانسته لباس یقه هفت را بدون وابستگی به لباس اولیه تولید کند. با این حال، روش +VITON-HD پیشنهادی یقه هفت را با دقت بالاتری نسبت به VITON-HD تولید کرده است. به عبارت دیگر، روش +VITON-HD پیشنهادی در مقایسه با دیگر روش‌ها، نتایج بهتری ارائه می‌دهد زیرا ویژگی‌های اصلی لباس هدف را به خوبی منتقل می‌کند. در حالی که روش‌های دیگر مثل CP-VTON و VITON-HD در مواجهه با لباس‌های اولیه با ویژگی های مختلف مشکلاتی دارند که باعث تغییرات ناخواسته در قسمت‌هایی مثل یقه می‌شوند. نتایج حاصل نشان می‌دهد که روش VITON-

+HD نه تنها در حفظ ویژگی‌های بصری لباس هدف و بهبود کیفیت تصاویر پُرو نهایی مؤثر است، بلکه قابلیت تعمیم‌پذیری به موقعیت‌های جدید را نیز دارد. این ویژگی‌ها، روش پیشنهادی را به یک رویکرد کارآمد و دقیق تبدیل می‌کند.

**نتیجه‌گیری**

در این پژوهش، ما یک شبکه پُرو مجازی مبتنی بر تصویر جدید به نام +VITON-HD را پیشنهاد کرده‌ایم که نتایج واقع‌گرایانه‌ای از پُرو مجازی لباس تولید می‌کند. با بهره‌گیری از ماژول پیش‌بینی نقاط‌کلیدی لباس به عنوان ورودی ماژول دفرمه‌سازی، به لباس‌های دفرمه دقیق‌تر و واقعی‌تری دست می‌یابیم که بدون نیاز به نمایش مستقل از لباس شخص، به‌درستی دفرمه

می‌شوند. این دقت بالاتر، منجر به بهبود نتایج پُرونهایی شده است. با توجه به لباس دفرمه، نقشه قطعه‌بندی بدن شخص در حالی که لباس هدف را بر تن دارد، پیش‌بینی می‌شود. در نهایت، با استفاده از روش نرمال‌سازی بخش‌های مرتبط‌نشده، اطلاعات گمراه‌کننده حذف شده و جزئیات لباس هدف حفظ می‌شود و تصویر پُرو نهایی توسط ژنراتور تولید می شود. مقایسه‌های کیفی نشان می‌دهند که این رویکرد، با حفظ کامل ویژگی‌های لباس هدف از جمله شکل و بافت کلی، به‌طور قابل‌توجهی از روش‌های پُرو مجازی موجود فراتر می‌رود. علاوه بر این، این روش پیشنهادی دارای قابلیت تعمیم در موقعیت‌های مختلف می‌باشد، که آن را به یک راهکار بسیار مؤثر و کارآمد تبدیل می‌کند. در آینده، شبکه‌های پیشرفته پُرو مجازی قادر خواهند بود با استفاده از یک تصویر دوبعدی، نقشه‌ای دقیق از بدن کاربران ایجاد کنند و با بهره‌گیری از فناوری‌های شبیه‌سازی پیشرفته و الگوریتم‌های دفرمه‌سازی، لباس‌هایی با اندازه‌های مختلف را به‌طور دقیق تطابق دهند. این شبکه‌ها امکان نمایش 360 درجه از کاربر را فراهم کرده و حرکت و هماهنگی لباس در زوایای گوناگون را شبیه‌سازی می‌کنند. این فناوری می‌تواند تجربه خرید آنلاین را با بهینه‌سازی فرآیند و ارائه حسی واقعی و حرفه‌ای به سطحی نوین ارتقا دهد.

## مراجع